\documentclass{article}

   \usepackage[preprint]{neurips_2025}

\usepackage[utf8]{inputenc} % allow utf-8 input
\usepackage[T1]{fontenc}    % use 8-bit T1 fonts
\usepackage{hyperref}       % hyperlinks
\usepackage{url}            % simple URL typesetting
\usepackage{booktabs}       % professional-quality tables
\usepackage{amsfonts}       % blackboard math symbols
\usepackage{nicefrac}       % compact symbols for 1/2, etc.
\usepackage{microtype}      % microtypography
\usepackage{xcolor}         % colors
\usepackage{graphicx}
\usepackage{array} % newly added
\usepackage{colortbl}
\usepackage{bookmark}
\usepackage{wrapfig}
\usepackage{lipsum}    % newly added
\usepackage{amsmath}
\usepackage{colortbl}
\usepackage{natbib}
\usepackage[export]{adjustbox}
\usepackage{float}

\newfloat{figtab}{htb}{fgtb}
\makeatletter
  \newcommand\figcaption{\def\@captype{figure}\caption}
  \newcommand\tabcaption{\def\@captype{table}\caption}
\makeatother

\setcitestyle{numbers,square}
\definecolor{mygray}{gray}{.9}

\title{From Next-Token to Next-Block: A Principled Adaptation Path for Diffusion LLMs}

\author{Peking University \& Huawei Technologies    }

\begin{document}

\maketitle

\begin{abstract}

  Diffusion Language Models (DLMs) enable fast generation, yet training large DLMs from scratch is costly. As a practical shortcut, adapting off-the-shelf Auto-Regressive (AR) model weights into a DLM could quickly equip the DLM with strong long-context generation capabilies. Prior “adaptation” attempts either modify logits or randomly grow attention masks to Full-Sequence diffusion, or simply transplant AR weights into a Block-Diffusion recipe, leaving two key questions unaddressed: where is the final destination of adaptation, and how to adapt better? For manifold benefits, we reframe the whole AR-to-DLM adaptation under the Block-Diffusion paradigm, transitioning from block size 1 to the final Block-Diffusion state. Concretely, the principled pathway of adaptation is designed as follows: we keep a context-causal path where causal attention is kept in the prefix, an efficient parallel adaptation procedure where an AR guidance is maintained, and gradual increment of the generation block size for a smoother transition. Built on these components, the adaptation is proved competitive on various models at different scales. With better adaptation, we propose \textsc{NBDiff-7B} that could inherit the long-context modeling and reasoning capabilities, and achieve state-of-the-art performance among the 7B-class DLMs. Codes: \url{https://github.com/YuchuanTian/NBDiff}.

\end{abstract}

\section{Introduction}
\label{sec:intro}

Large language models (LLMs) are rapidly permeating real-world applications because of their strong generative capability. However, the dominance of AutoRegressive (AR) LLMs is built on a fundamental trade-off: powerful left-to-right causal generation at the cost of strictly sequential, token-by-token decoding. This trade-off creates an inference bottleneck that limits the decoding speed of AR LLMs. In contrast, Diffusion Language Models (DLMs) offer a promising alternative by enabling parallel generation, reducing sequential dependencies and yielding substantially higher throughput and lower wall-clock latency in practice.

Current diffusion approaches for language have largely converged on masked diffusion, with two dominant paradigms. Full-Sequence Diffusion~\cite{llada,dream} starts from a fully masked sequence and denoises to a complete output where all tokens attend bidirectionally. Block-Diffusion~\cite{blockdiff,sdar} decodes one block at a time: tokens are bidirectional within the active block, while blocks themselves follow left-to-right causal order, yielding a semi-autoregressive workflow. Training masked diffusion is intrinsically harder than AR pretraining because, unlike AR--where every token contributes a next-token-prediction loss--only masked tokens provide supervision, which slows optimization. Yet masked diffusion and AR models are strikingly similar in input–output format and transformer architecture. This naturally motivates the question: can we leverage powerful off-the-shelf AR checkpoints and rapidly adapt them into diffusion models, preserving their knowledge while avoiding the cost of training a DLM from scratch?

Existing adaptation methods are lacking. Early attempts used logit shifts and random attention mask growth to Full-Sequence Diffusion~\cite{adapt,dream}. More recent block-wise adaptations simply 'transplant' the AR model into a Block-Diffusion training setup 'as is.'~\cite{sdar}--they do not investigate the core mismatch between AR and Block-Diffusion. These methods leaves a clear gap: where should be our destination of adaptation, and how to adapt an AR model to Diffusion for better performance?

\begin{wrapfigure}{r}{0.5\textwidth}
    \centering
    \vspace{-10pt}
    \includegraphics[width=0.5\textwidth]{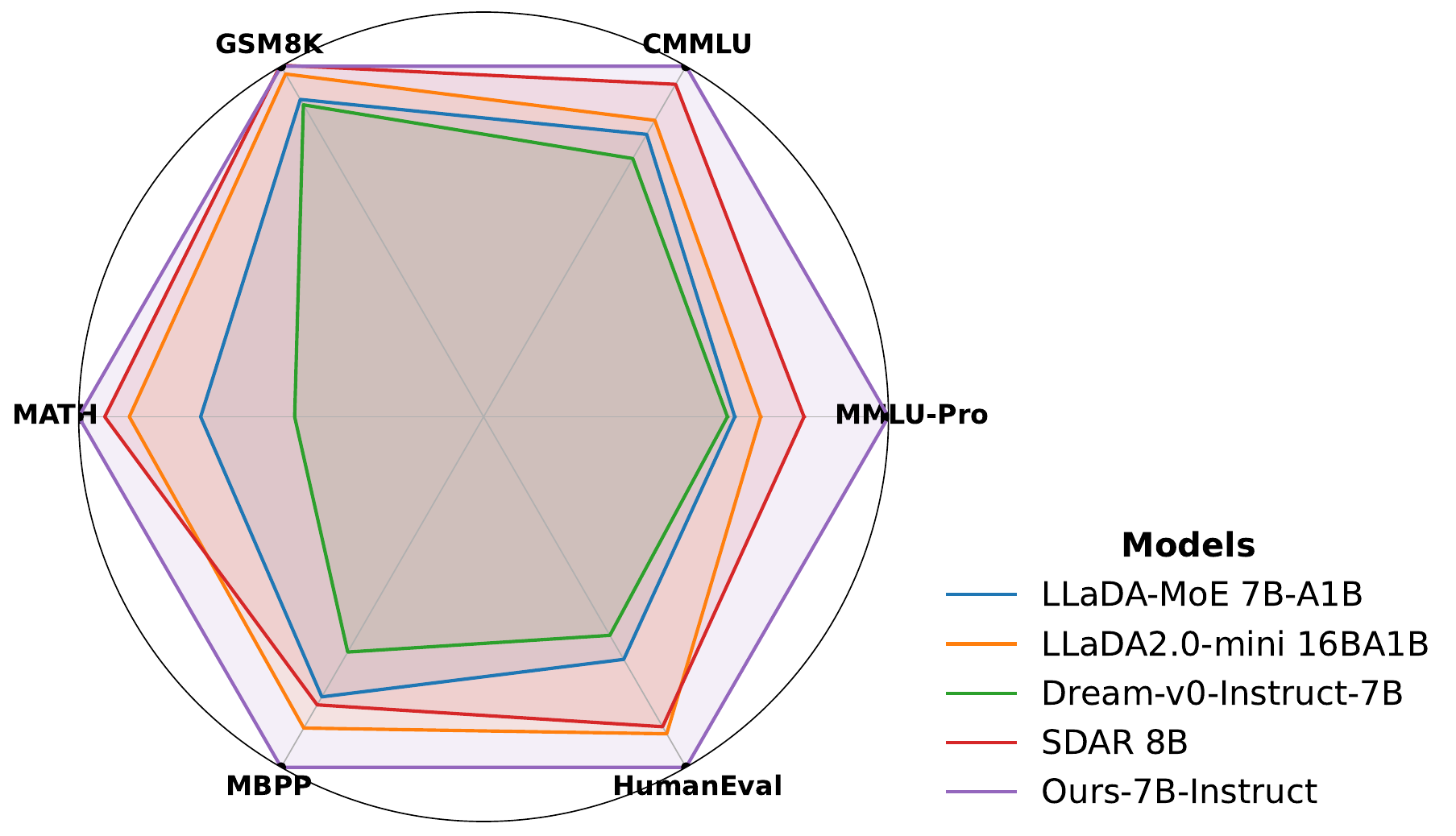}
    \caption{\textbf{Comparison of our model with baselines.} After adaptation from an open-sourced AR LLM, our model has good long-sequence and reasoning capabilities and shows outstanding performance in various benchmarks.}
    \vspace{-10pt}
    \label{fig:radar}
\end{wrapfigure}

Our approach is grounded in a key insight: for longer-sequence training, efficient inference, and adaptation benefits, Block-Diffusion should be either the pathway and the destination of adaptation. AR generation could be viewed as a special case of Block-Diffusion with a $blocksize$ of 1, reframing adaptation not as a crude switch, but as a smooth transition across a spectrum. Under this unified view, we look for a principled and smooth transition path from AR to Block-Diffusion. Our design consists of a context-causal attention mask that preserves AR inductive bias in committed context, parallel training with an auxiliary AR guidance that regularizes the path of adaptation, and gradual growth of block size. The design provides an efficient adaptation strategy from AR to DLM that progressively unlocks bidirectional attention and parallel decoding within the generating block while maintaining strict train–inference alignment.

Our contributions are as follows:

\begin{enumerate}
  \item After investigation, we propose to view the whole adaptation process under Block-Diffusion for its natural training, inference, and adaptation benefits. The transition from AR to DLM is then simply blocksize growth from causal ($blocksize=1$) to target size.
  \item We propose the Context-Causal mechanism tailored for this adaptation, which preserves AR knowledge in the context while enabling efficient bidirectional intra-block generation. We develop an efficient parallel training strategy that aligns with inference and incorporates an auxiliary AR loss. We also develop a gradual block growth approach that alleviates the gap between AR and Block-Diffusion models. These measures markedly improve adaptation performance.
  \item We demonstrate the effectiveness of our approach with various models. We also propose \textsc{NBDiff-7B}, which, after efficient adaptation from its strong AR counterparts, could model \textbf{long contexts} (up to 32K sequence lengths) and perform \textbf{reasoning}. Both \textsc{NBDiff-7B-Base} and \textsc{NBDiff-7B-Instruct} outperform strong baselines like LLaDA~\cite{llada,lladamoe,llada20}, Dream~\cite{dream}, and SDAR~\cite{sdar} on general, math, and code benchmarks (Figure~\ref{fig:radar}), achieving state-of-the-art performance.
\end{enumerate}

\begin{figure}[htbp]
    \centering
    \includegraphics[width=0.9\textwidth]{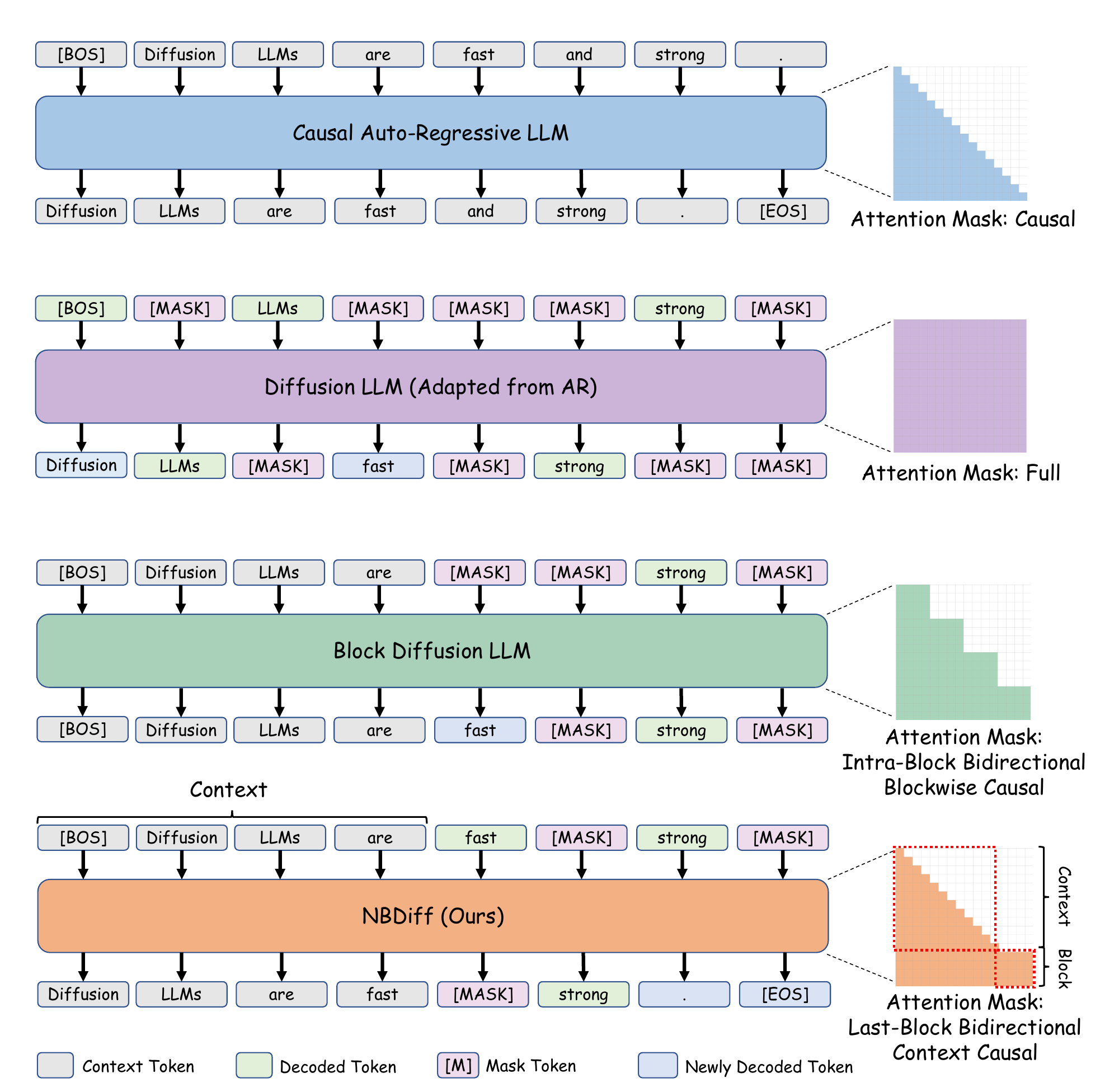}
    \vspace{-10pt}
    \caption{\textbf{The diffusion paradigm of our \textsc{NBDiff-7B} model.} We compare popular language generation paradigms. Diffusion LLMs adapted from AR adopt logit shift and attention mask growth; Block-Diffusion uses block-wise autoregressive and maintains an intra-block bidirectional mask; Our model adopts Block-Diffusion where bidirectional attention is used intra-block, but features a causal context.}
    \label{fig:plot1}
    \vspace{5pt}
  \end{figure}

\section{Related Work}
\label{sec:relatedwork}

\paragraph{Discrete diffusion language models.}
Diffusion has been extended to categorical spaces, showing that discrete denoising objectives can effectively model text and clarifying connections to classical LM training, including transition design and absorbing states \cite{Austin2021D3PM}. Two dominant paradigms have since emerged. Masked diffusion iteratively reveals tokens, enabling controllable generation with competitive likelihoods without left-to-right decoding \cite{Li2022DiffusionLM}, while recent MDLM variants substantially close the perplexity gap to AR LMs using simplified training recipes \cite{Sahoo2024MDLM}. Absorbing-state diffusion instead corrupts tokens toward a sink symbol; recent analyses relate its objective to conditional modeling, calibration, and sampling behavior \cite{Ou2024RADD}. At scale, systems trained from scratch such as LLaDA demonstrate that masked-diffusion-style pretraining can rival strong AR baselines and extend naturally to multimodal instruction tuning \cite{llada,You2025LLaDAV}, establishing the viability of DLMs at billion-parameter scale.

\paragraph{Recent trend: Block-Diffusion, adaptation from AR, and test-time scaling.}
While Full-Sequence Diffusion provides fully bidirectional context, it is computationally inefficient for long texts and misaligned with left-to-right inductive biases. Attempts to amortize this cost via intermediate-state caching improve efficiency but do not fundamentally resolve the issue \cite{dllmcache,dkvcache,fastdllm}. Block-Diffusion addresses this by fixing past context while denoising the current block bidirectionally, enabling parallel token updates and unbounded-length generation with tunable quality--efficiency trade-offs \cite{Han2023SSDLm,blockdiff}. Beyond training from scratch, several works adapt pretrained AR models into diffusion-style decoders, often at the block level, reporting objective connections, practical conversion recipes, and hybrid AR--diffusion paradigms that preserve AR quality while enabling parallel generation \cite{adapt,sdar,fastdllm,fastdllmv2}. In parallel, diffusion-based reasoning systems explore inference-time scaling or reinforcement learning to improve multi-step reasoning, particularly for math and code, but remain limited by short contexts or underutilized AR priors \cite{diffusionthoughts,remask,thinkmask,diffucoder,d1,wd1,d2}. In contrast, our approach adapts strong AR models into block-diffusion generators via a smooth way, enabling longer context and better performance.

\section{Rethinking DLM Adaptation from AR: to Where, and How?}

\subsection{Revisiting Previous Adaptation}

Prior adaptation work~\cite{adapt} mainly focuses on adapting a Full-Sequence Diffusion model from an AR model. The authors observe the difference in attention mechanism, and proposes random "annealing", or random growth, of the attention mask from a lower-triangular causal attention mask to a full, bidirectional attention mask.

While the work~\cite{adapt} is trying to bridge the AR and Diffusion generation paradigms, we hold different opinions on random growth of attention masks: its transition is not "natural." In practice, training sees unknown future corpora; sporadically granting early tokens access to a random subset of future tokens yields incomplete and potentially misleading context, thus limiting adaptation potentials. 

Hence, we try to answer two major questions regarding this transition, \textit{i.e.} to \textbf{where}, and \textbf{how}. Firstly, what should be the destination of this transition? Secondly, is there a smoother and better way to transition from an AR model to a DLM model?

\subsection{Block-Diffusion and its Advantages}

Unlike previous adaptation methods~\cite{adapt} that focuses merely on Full-Sequence Diffusion models, recent diffusion LLMs~\cite{sdar,fastdllmv2} increasingly adopt Block-Diffusion~\cite{blockdiff}, which is both more efficient and performant than Full-Sequence Diffusion and conceptually sits between Full-Sequence Diffusion and AR: generation proceeds left-to-right across blocks while remaining bidirectional within a block (Figure~\ref{fig:plot1}). Specifically, tokens within the same block attend to each other bidirectionally, whereas attention across blocks is strictly causal. Decoding is performed block by block, with all tokens in a block capable of generating in parallel.

We analyze the advantages from several aspects: either from training, from inference, and also from the perspective of adaptation from AR models.

\begin{figure}[htbp]
  \centering
  \begin{minipage}[t]{0.48\textwidth}
    \centering
    \includegraphics[width=\linewidth]{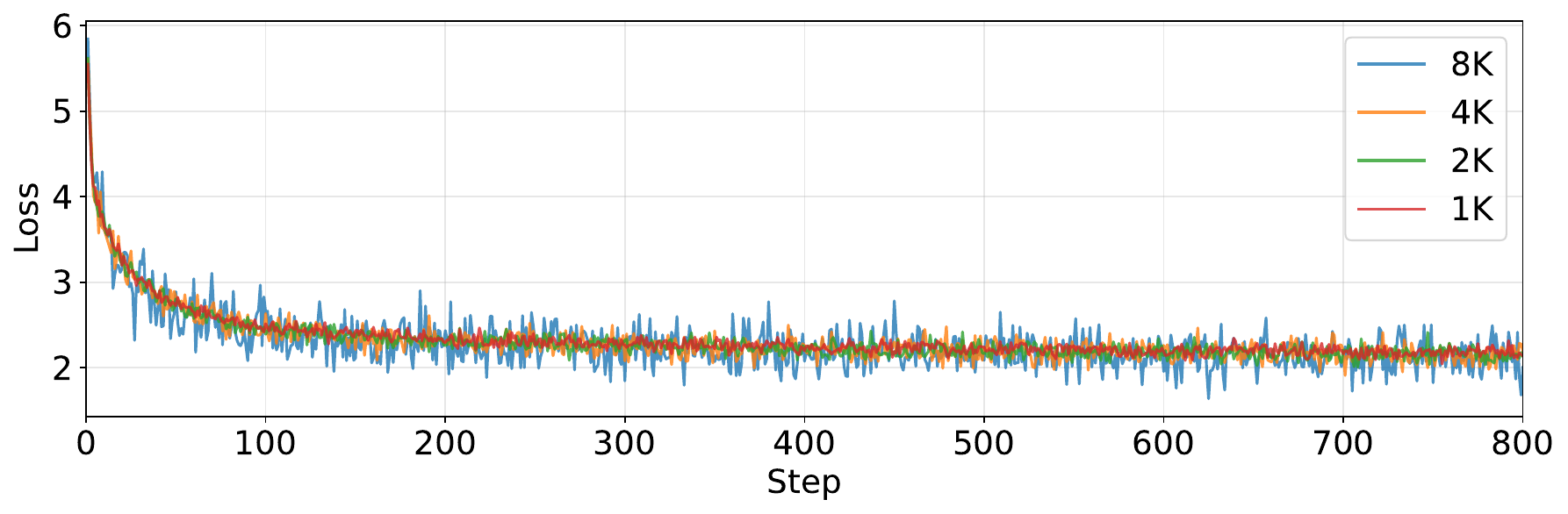}
  \end{minipage}\hfill
  \begin{minipage}[t]{0.48\textwidth}
    \centering
    \includegraphics[width=\linewidth]{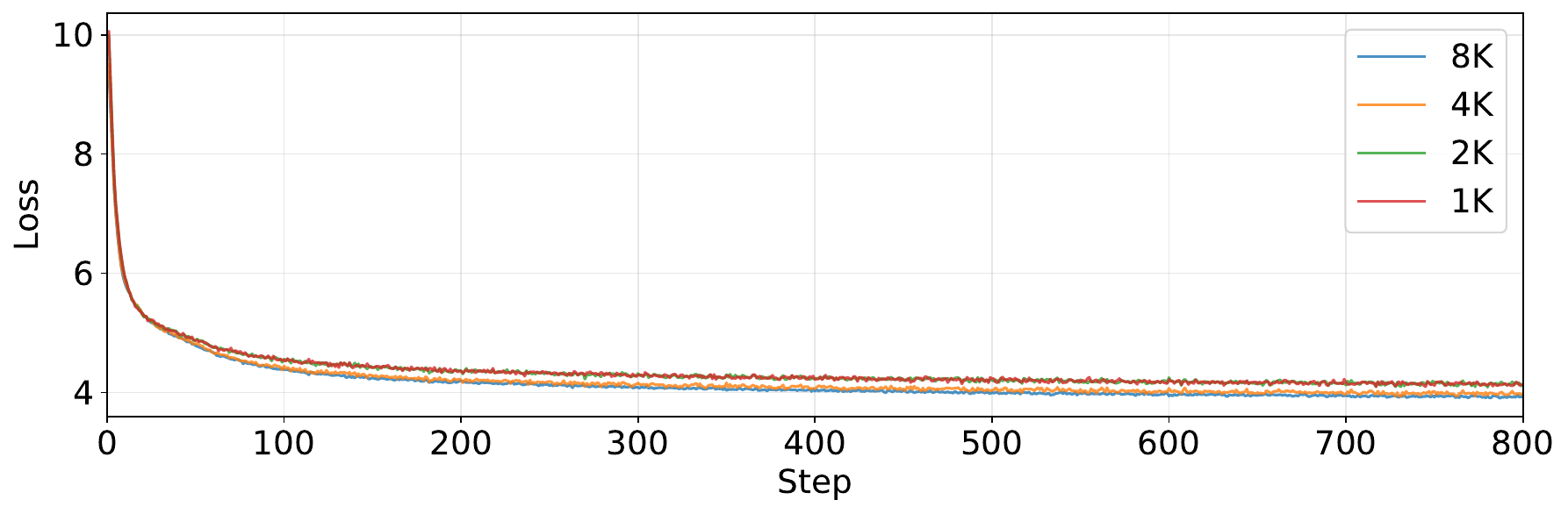}
  \end{minipage}
  \caption{\textbf{Loss curves at different sequence lengths (Left: Full-Sequence; Right: Block-Diffusion).} As sequence length increases, Full-Sequence Diffusion suffers loss fluctuation while Block-Diffusion loss remains stable. Block-Diffusion is a better choice for long-sequence generation.}  
  \label{fig:seq_loss}
\end{figure}

\paragraph{Training advantages: stabler training at longer sequences.} AR models excel at long-sequence reasoning, yet most diffusion LLMs still operate at modest context lengths (e.g., LLaDA~\cite{llada}=1K, Dream~\cite{dream}=512), raising the question of how sequence length impacts DLM training. We therefore pretrain Full-Sequence Masked Diffusion and Block-Diffusion under identical corpora at 1K/2K/4K/8K sequence lengths and compare their training losses (Figure~\ref{fig:seq_loss}). For fair comparison, we adjust batchsize accordingly with sequence length to ensure that each training iteration consumes the same number of tokens. As context grows, the full-sequence model exhibits increasingly large loss oscillations, whereas Block-Diffusion remains consistently stable. Notably, at long lengths the block-diffusion loss is also lower, suggesting that longer contexts provide tangible generation benefits when the training dynamics are well-conditioned.

We hypothesize the instability of masked Full-Sequence Diffusion stems from the combinatorial explosion of the masking space: as sequence length increases, (i) \textit{the number of masked tokens per step can vary drastically}, and (ii) \textit{the effective decoding/denoising orders proliferate, complicating optimization}~\cite{kim2025trainworstplanbest}. In contrast, Block-Diffusion constrains the space by fixing a constant block size, which regularizes the denoising schedule, preserves left-to-right structure, and stabilizes gradients, thereby enabling long-sequence advantages to manifest in masked DLMs.

\paragraph{Inference advantages: KV-Cache reuse.} Different from Full-Sequence Diffusion where the whole sequence has to pass through the model together for each inference step, Block-Diffusion keeps previous tokens fixed while only performs decoding in the last block of the generated tokens. Thus it could re-use the KV-Cache from previous block generations and only the last block to be generated needs to be passed into the model, reducing significant inference costs. Besides, though Block-Diffusion has designated the causal (left-to-right) block generation sequence, the use of bidirectional attention within the last generating block still enables parallel token-decoding. In practice, we use the $blocksize$ of 32 tokens (larger than previous same-scale DLMs~\cite{sdar}) to tap the speed potential of the proposed model to the full.

\paragraph{Adaptation advantages: analogous paradigms and easier adaptation from AR.} Apart from the performance advantages, Block-Diffusion helps easy adaptation from AR. Instead of forcing a global jump from causal to full-sequence bidirectional attention, we treat Block-Diffusion as the destination. By preserving left-to-right semantics at the block level and relaxing bidirectionality inside the active block, we try to partially align with the AR inductive bias for better adaptation, which greatly reduces the difficulty for alignment. In addition, the blockwise semi-AR manner could also enable parallel training, improving data utility and model convergence.

In the next section, we will stick to the paradigm of Block-Diffusion and seek a way for fast transition from AR model.

\section{Designing Transition Paradigms}
\label{sec:method}

\subsection{How Should the Unmasked Context Attend?}
Comparing the attention mechanisms of Block-Diffusion and  AR (which could be roughly viewed as Block-Diffusion of $blocksize=1$, as introduced in Sec.~\ref{sec:intro}), the key difference that requires our adaptation efforts lies in the bidirectional attention within the last active block; namely, we have to grow the attention mask at the end of the generating sequence from $blocksize=1$ to target block size. However, apart from the attention within the active block region, different transition solutions arises from the decoded context: how tokens in the unmasked context ($x_{<s_K}$) should attend to each other? Here we analyze two possible attention pathways of Block-Diffusion as follows:
\begin{itemize}
\item \textbf{Block-Causal (widely used in Block-Diffusion~\cite{blockdiff} / D2F~\cite{d2f} / SDAR~\cite{sdar}).}
Tokens have \emph{bidirectional} attention {within every block} (both past/committed blocks and the active block), and \emph{causal} flow {across blocks} (each token can see all tokens in earlier blocks). This maximizes intra-block interaction everywhere, not only in the active block.

\item \textbf{Context-Causal (our preferred setting).}
The {context} (prompt $+$ already generated/committed blocks) remains \emph{strictly causal}: each token only attends to itself and predecessors.
{Only the last (active) block} is given \emph{bidirectional} attention to support diffusion-style refinement; future blocks are hidden.
\end{itemize}

To examine the two schemes, we adapt two Block-Diffusion models from an AR model (based on Block-Causal and Context-Causal, respectively) by training 2000 iterations, and examine their performance on popular math and coding benchmarks. The results are shown in Table~\ref{tab:scheme}.

\begin{table}[htbp]
    \centering
    \caption{\textbf{Comparison of Block-Causal and Context-Causal attention schemes.} Context-Causal gains a clear advantage in adaptation from AR.}
    \scalebox{0.9}{%
    \setlength{\tabcolsep}{2pt}
    \begin{tabular}{lccccc}
    \toprule
    Scheme & GSM8K & MATH & HumanEval & MBPP & Avg \\
    \midrule
    \textbf{Block-Causal}   & {60.1} & {1.6} & {24.4} & {39.4} & {31.4} \\
    \rowcolor{gray!10}\textbf{Context-Causal} & \textbf{{68.8}} & \textbf{{36.8}} & \textbf{{41.5}} & \textbf{{47.4}} & \textbf{{48.6}} \\
    \bottomrule
    \end{tabular}
    }
    \vspace{2mm}
    \label{tab:scheme}
\end{table}

\paragraph{Empirical takeaway and intuition.}
In these preliminary adaptation experiments, \textbf{Context-Causal} consistently outperforms Block-Causal by large margins: the accuracy is {significantly higher} when the context keeps strict causality and only the active block is bidirectional.

We attribute this to:
(i) \emph{Inductive-bias alignment} with AR pretraining, which reduces the gap between AR and Block-Diffusion by preserving causal self-attention in the context; and
(ii) \emph{Generation-paradigm consistency}: although the active block is refined bidirectionally (no fixed order inside the block), the overall decoding remains left-to-right \emph{across} blocks.
Keeping the context causal does not reduce the visibility required for the block being generated and avoids introducing spurious, partially bidirectional signals into earlier (already ``finalized'') content.

\begin{figure}[htbp]
    \centering
    \includegraphics[width=0.9\textwidth]{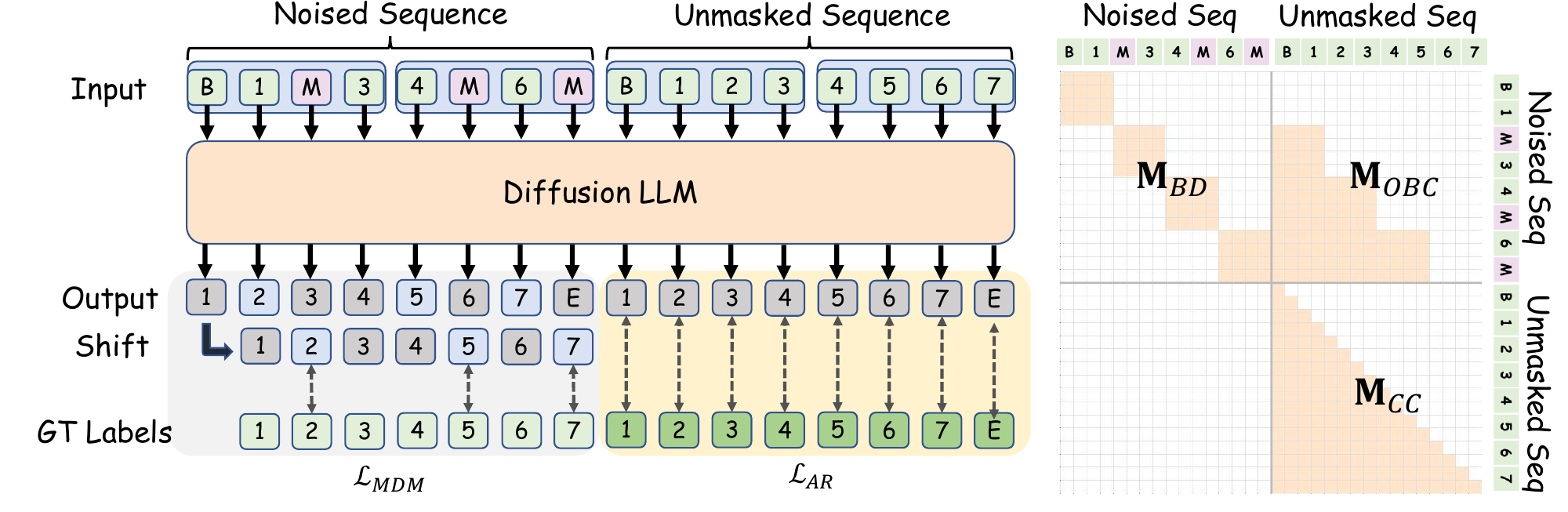}
    \vspace{-10pt}
    \caption{\textbf{Our Parallel Training Diagram.} The diagram shows the parallel training form of our Context-Causal setting (we use $blocksize=4$ as an example; the actual $blocksize$ is 32). We concatenate a clean, unmasked token sequence to the noised sequence. The attention mask $\mathbf{M}_{\mathrm{all}}$ is designed (shown in the right) such that strictly-causal attention is applied in the unmasked input; for the masked input, each token has bidirectional attention intra-block, but causal attention to past inter-block tokens that are unmasked. AR loss $\mathcal{L}_{\mathrm{AR}}$ is introduced in addition to the canonical masked loss $\mathcal{L}_{\mathrm{MDM}}$ for faster adaptation.}
    \label{fig:plot2}
    \vspace{-1pt}
  \end{figure}

\subsection{Training Parallelism}

The naive block-diffusion recipe is data-inefficient: random cropping wastes the remaining tokens of each sequence, and only a small subset of masked tokens inside the last block contributes to the loss. Unlike AR pretraining—where every token can supervise next-token prediction—switching to next-{block} prediction sharply reduces token utilization. We therefore restructure training so that all blocks provide learning signal in a single forward pass. 

To reach this goal, the clean (unmasked) sequence is concatenated after the noised sequence and enforce a structured attention mask that lets the clean side provide context to the noised side. The mask (shown in Figure~\ref{fig:plot2}) has four quadrants: (upper-left, $\mathbf{M}_{\mathrm{BD}}$) block-diagonal self-attention within the noised view (bidirectional inside each noised block); (upper-right, $\mathbf{M}_{\mathrm{OBC}}$) attention from noised tokens to earlier clean blocks, so denoising conditions on stable context; (lower-left) zeros, preventing the clean sequence from reading the noised view (matching inference); and (lower-right, $\mathbf{M}_{\mathrm{CC}}$) strictly causal self-attention within the clean sequence. Relative to prior block-diffusion training masks~\cite{blockdiff}, we replace block-causal with fully causal (context-causal) in this lower-right quadrant, preserving the AR inductive bias in the context while still enabling intra-block bidirectionality only where generation happens. The detailed formulation of training and the structured attention mask in enclosed in Appendix~\ref{sec:methodologydetails}. This design optimizes per-step token utilization, and empirically stabilizes training.

\subsection{AR Loss Guidance}
While our one-pass parallel recipe enables efficient Block-Diffusion training, the diffusion loss is only applied to logits on the noised (active) blocks; logits on the clean-context branch of the concatenated sequence mainly serve as conditioning signals. Meanwhile, our adaptation follows a path from Block-Diffusion with $blocksize=1$ (i.e., AR) to a target block size, and we would like this path to remain anchored to the AR behavior rather than drifting too far away. To this end, we introduce an auxiliary AR objective as a lightweight constraint, which is naturally attached to the clean-context branch because it already follows strictly causal self-attention. This turns otherwise unused context predictions into supervised next-token targets, improving token utilization without changing diffusion-side conditioning.

Let $\mathcal{C}$ index tokens on the clean context, $x_i$ be the ground-truth token at position $i$, and $\mathbf{M}_{\mathrm{CC}}$ denote the context-causal mask; we attach a standard LM head and define an autoregressive objective over the context as
\begin{equation}
\label{eq:ar_loss}
\mathcal{L}_{\mathrm{AR}}(\theta)=\mathbb{E}\Bigg[-\sum_{i\in\mathcal{C}}\log p_{\theta}\!\big(x_{i+1}\mid x_{\le i};\,\mathbf{M}_{\mathrm{CC}}\big)\Bigg].
\end{equation}
Let $\mathcal{L}_{\mathrm{MDM}}(\theta)$ be the masked/block-diffusion denoising loss computed on the noised view under $\mathbf{M}_{\mathrm{all}}$ (as in Eq.~\eqref{eq:parallel_loss}); we then train with an affine combination controlled by $\lambda\!\ge\!0$:
\begin{equation}
\label{eq:total_loss}
\mathcal{L}_{\mathrm{total}}(\theta)=\mathcal{L}_{\mathrm{MDM}}(\theta)+\lambda\,\mathcal{L}_{\mathrm{AR}}(\theta).
\end{equation}
In practice, we set $\lambda=0.5$ so that the number of tokens participating in the MDM and AR losses is kept at a comparable scale throughout adaptation. Overall, $\mathcal{L}_{\mathrm{AR}}$ provides a simple yet effective guidance signal along the AR$\rightarrow$Block-Diffusion path, while being ``free'' to compute from the clean-context branch in our parallel training formulation.

\subsection{Gradual Block Growth}

\begin{figure}[htbp]
    \centering
    \includegraphics[width=0.9\textwidth]{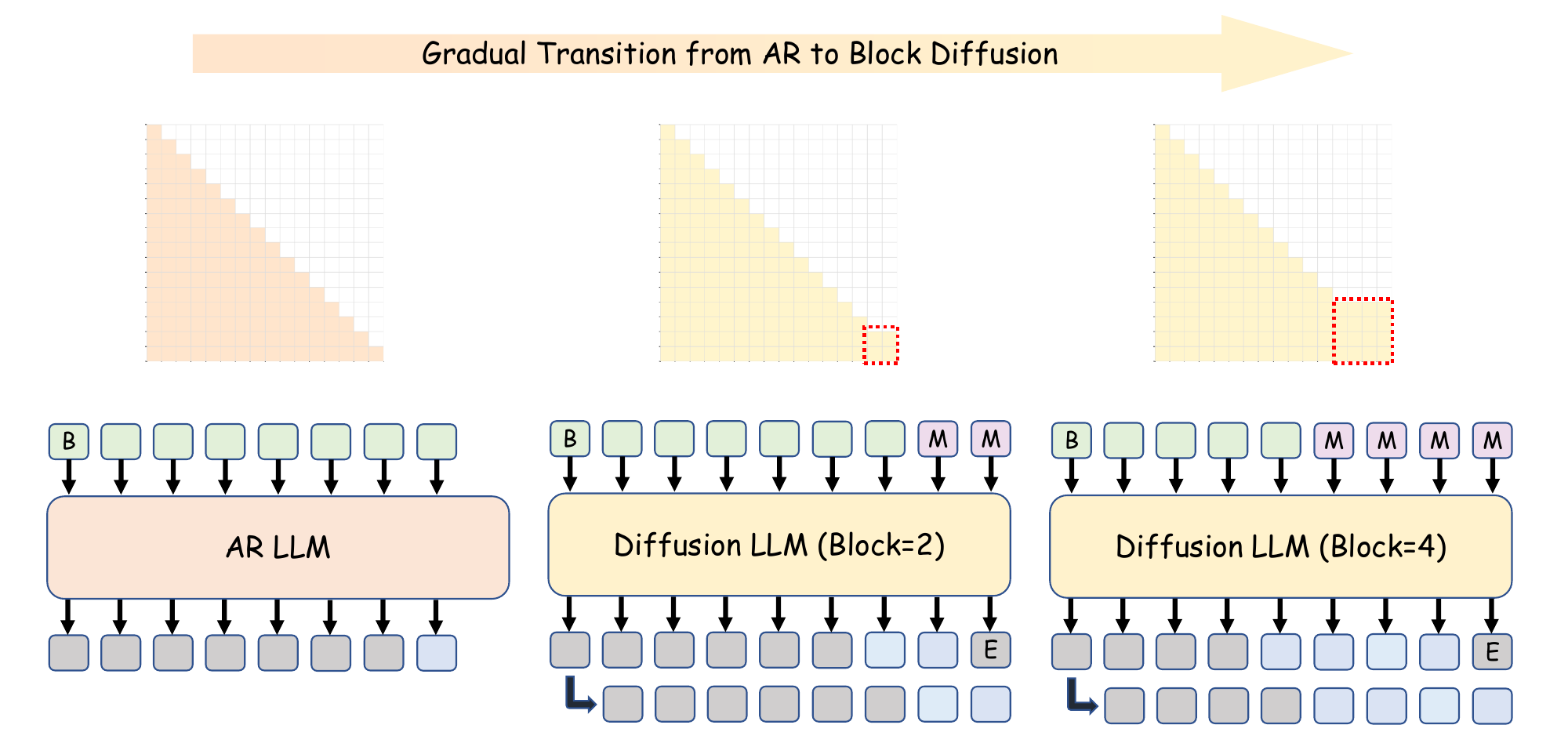}
    \vspace{-10pt}
    \caption{\textbf{The Diagram for Gradual Block Growth.} Starting from an AR model (which could be viewed as a Block-Diffusion model of $blocksize=1$), we gradually double the $blocksize$ during training until reaching the target size, mitigating the adaptation gap between AR and Block-Diffusion.}
    \label{fig:plot3}
    \vspace{-1pt}
  \end{figure}

After determining the path for transition and guidance, we pursue a smoother adaptation by growing blocks in Block-Diffusion (Figure~\ref{fig:plot3}). As noted, AR could be viewed as Block-Diffusion of $blocksize=1$; hence, the transition could be viewed under the Block-Diffusion paradigm, where we start from block size of 1 and end at the target block size. Naturally, we gradually increase the generation block size from AR's single-token steps toward larger blocks, so that the model transitions continuously from next-token prediction to next-block refinement. This monotonic growth retains left-to-right causality while progressively introducing intra-block bidirectionality, narrowing the procedural gap and easing optimization.

Starting from blocksize $b{=}1$ (AR), we interpolate to larger bidirectional blocks by growing $b$ in \emph{integer powers} of a user-chosen base $r\in\{2,4,\dots\}$ (on a normal basis the power of 2 is adopted) at fixed training intervals. Let $s$ be the global training step, $\Delta$ the interval (in steps) between growth events, $s_0$ an optional warmup before the first growth, $b_0{=}1$ the starting size, and $b_{\max}$ the inference target. The schedule is as follows.
\begin{equation}
\label{eq:int_power_growth}
b(s)\;=\;\min\!\Big\{\,b_{\max},\; b_0 \cdot r^{\,\big\lfloor \tfrac{\max(0,\,s-s_0)}{\Delta}\big\rfloor}\Big\},
\end{equation}
which holds $b$ constant on plateaus $[s_0+k\Delta,\; s_0+(k{+}1)\Delta)$ and multiplies it by $r$ whenever $s$ crosses a multiple of $\Delta$ (e.g., $1\!\to\!r\!\to\!r^2\!\to\!\dots$) until capped by $b_{\max}$. This integer-power curriculum reduces the AR $\to$ diffusion adaptation gap by aligning early optimization with the AR inductive bias (small $b$) and gradually unlocking intra-block bidirectionality and parallel supervision as $b$ increases. In practice, we keep train-inference semantics matched at each plateau via the same context-causal mask family, optionally co-schedule compute by shrinking the refinement steps per block $T(s)\propto 1/b(s)$ to maintain a roughly stable token-update budget, and anneal the AR-loss weight $\lambda$ as blocks grow to reallocate gradient capacity toward the diffusion objective.

\begin{table}[!t]
    \centering
    \caption{\textbf{Comparison of Adaptation Methods.} We demonstrate the effectiveness of our method across different models and settings.We also show the contribution of each component in our adaptation methods.}
    \scalebox{0.9}{%
    \begin{tabular}{lccccc}
    \toprule
    Method & GSM8K & MATH & HumanEval & MBPP & Avg \\
    \midrule
        \textit{Qwen3-4B-Base} & & & & & \\
    Annealed Attention Mask~\cite{adapt} & {76.57} & {28.66} & {25.61} & {59.40} & {47.56} \\
    Plain Finetuning & {72.18} & {24.84} & \textbf{31.10} & {56.80} & {46.23} \\
    \rowcolor{gray!10}\textbf{Ours} & \textbf{79.83} & \textbf{32.26} & {27.44} & \textbf{61.60} & \textbf{50.28} \\\midrule    
        \textit{Qwen3-8B-Base} & & & & & \\
    Annealed Attention Mask~\cite{adapt} & {80.67} & {25.76} & {39.02} & {49.80} & {48.81} \\
    Plain Finetuning & {78.32} & {24.22} & \textbf{42.68} & {58.20} & {50.85} \\
    \rowcolor{gray!10}\textbf{Ours} & \textbf{82.26} & \textbf{34.68} & {31.71} & \textbf{64.40} & \textbf{53.26} \\\midrule
        \textit{openPangu-7B} & & & & & \\
    Annealed Attention Mask~\cite{adapt} & {70.05} & {35.46} & {26.83} & {45.00} & {44.34} \\
    Plain Finetuning & {72.63} & {36.14} & {39.63} & {47.40} & {48.95} \\
    {+ AR loss} & {75.13} & {43.30} & {40.24} & {53.20} & {52.97} \\
    \rowcolor{gray!10}{+ AR loss \& Gradual Growth }(\textbf{Ours}) & \textbf{76.57} & \textbf{44.06} & \textbf{44.51} & \textbf{54.60} & \textbf{54.94} \\
    \bottomrule
    \end{tabular}
    }
    \vspace{2mm}
    \label{tab:alladapt}
\end{table}

\section{Experiments}
\label{sec:experiments}

In this section, we empirically demonstrate the success of our proposed adaptation methods via manifold experiments. We test our method on various off-the-shelf model weights of different sizes, including Qwen3-4B-Base, Qwen3-8B-Base~\cite{qwen3}, and openPangu-Embedded-7B~\cite{panguembedded}.

\paragraph{Experiment setup.} For all experiments, we use sequence length $\ell{=}8\mathrm{k}$ and global batch size $B{=}1024$; $lr$ is set as $1e-5$ with the Adam optimizer. we train for 4000 iterations that consumes approximately 30B of the training data. Adaptation iteration for gradual growth is set as 2500. For the Qwen3 series, we use the FineWeb-100B~\cite{fineweb} dataset. For openPangu, we use a large, high-quality 700B internal dataset.

\paragraph{Comparison with existing baselines.} On the basis of logit shift introduced in DiffuLLaMA~\cite{adapt}, we adopt two methods as baselines. "Annealed Attention Mask"~\cite{adapt} proposes random growth from the auto-regressive causal attention mask to the targeted Diffusion attention mask. In our setting, since we are using structured attention masks for parallel training, "Annealed Attention Mask" is applied as the random interpolation between structured attention masks for $blocksize=1$ and $blocksize=32$. We are aware that $\mathbf{M}_{\mathrm{BD}}$ and $\mathbf{M}_{\mathrm{OBC}}$ are closely dependent; hence, we "chain" the randomness of $\mathbf{M}_{\mathrm{BD}}$ and $\mathbf{M}_{\mathrm{OBC}}$ such that each token will not view each position twice in attention. "Plain Finetuning" directly uses the targeted attention mask for training.

As shown in Table~\ref{tab:alladapt}, our method consistently outperforms both baselines across model scales and evaluation benchmarks. Compared to annealed attention masks and plain finetuning, our approach achieves the highest average performance on Qwen3-4B, Qwen3-8B, and openPangu-7B with particularly notable gains on GSM8K, MATH, and MBPP. These results indicate that a structured and progressive adaptation strategy is more effective than either random interpolation or directly applying the target attention mask, leading to more stable and reliable performance improvements across diverse tasks and model sizes.

\paragraph{Contribution of adaptation.} In Table~\ref{tab:alladapt}, we also ablate our two adaptation components: AR loss and gradual block-size growth—starting from a plain fine-tuning baseline. Adding AR loss lifts the overall Avg from 48.95 to 52.97 (+4\%), with the largest gains on math and MBPP and modest improvements elsewhere. Stacking gradual block-size growth further raises Avg to 54.94, indicating that smoother progression from next-token to next-block generation improves stability and yields consistent, additional benefits—especially for coding and multi-step reasoning.

\section{NBDiff-7B}
\label{sec:nbdiff}

We further demonstrate the effectiveness of our adaptation method via intensive data scaling. We start from the {openPangu-Embedded-7B}~\cite{panguembedded} base checkpoint and adapt it into a diffusion language model (DLM) using the training dataset in that release. With these efforts, we launch \textsc{NBDiff-7B}, a State-of-the-Art diffusion model.

\subsection{Setup}

\paragraph{Training.} The pretraining adaptation stage uses a two-phase learning-rate schedule: we keep the learning rate {constant} for the first $24{,}000$ iterations and then apply a {learning-rate cooldown} over the final $60{,}000$ iterations, for a total of $84{,}000$ iterations. We train with sequence length $\ell{=}8\mathrm{k}$ and global batch size $B{=}1024$. The effective tokens processed per iteration are 8M tokens, so across $84{,}000$ iterations the total token volume is approximately 700B tokens. \textsc{NBDiff-7B-Base}, the State-of-the-Art base model with 8K context, is completed after this stage. Then, to equip the model with long-sequence generative capabilities, we extend the pretraining sequence length $\ell{=}32\mathrm{k}$ and train for $23{,}800$ iterations (approximately 100B tokens), equipping the model with long-sequence modeling capabilities. Finally, we use 10B-token SFT data of sequence length $\ell{=}32\mathrm{k}$ to finetune the model for 10 epochs (approximately $17{,}000$ iterations) and equip it with reasoning capabilities.

We use a {uniform} masking strategy over the diffusion step $t\!\sim\!\mathrm{Uniform}[0,1]$ (sampled and mapped to the discrete step index), and keep the inference/training mask families matched at each curriculum plateau. All other optimizer and system-level settings follow the default configuration of the openPangu-Embedded-7B~\cite{panguembedded} release.

\paragraph{Inference.} Our inference sampling is mainly based on the implementation of Fast-DLLM-v2~\cite{fastdllmv2}: at the macro level the sequence is generated left-to-right by blocks of size 32 (causal across blocks), while inside each block we permit bidirectional attention to refine tokens jointly. For speed, the inner refinement can follow the v2 “small-block” schedule or be collapsed into a single full-block bidirectional pass when latency matters. For general benchmarks, we adopt the standard configuration. Greedy strategies are used for math tasks and $p=0.9, T=1$ sampling strategies are used for coding benchmarks to optimize performance. The experiment results are all single-run outcomes.

\begin{table}[h!]
    \centering
    \caption{\textbf{Comparison between \textsc{NBDiff-7B-Instruct} and the latest SFT (Instruct) version diffusion language models.} Our model demonstrates strong performance on general, math, and coding tasks, and outcompetes latest diffusion baselines by large margins. * indicates non-official replications.}
    \scalebox{0.9}{
    \begin{tabular}{lcccc>{\columncolor{gray!10}}c}
        \toprule
         & \multicolumn{1}{c}{\textbf{LLaDA-MoE}} & \multicolumn{1}{c}{\textbf{LLaDA2.0-mini}} & \multicolumn{1}{c}{\textbf{Dream-v0}} & \multicolumn{1}{c}{\textbf{SDAR}} & \multicolumn{1}{c}{\textbf{Ours-7B}} \\ 
        {Benchmark} & \multicolumn{1}{c}{\textbf{7B-A1B}} & \multicolumn{1}{c}{\textbf{preview 16BA1B}} & \multicolumn{1}{c}{\textbf{Instruct-7B}} & \multicolumn{1}{c}{\textbf{8B}} & \multicolumn{1}{c}{\textbf{Instruct}} \\ \midrule
        \textit{General} &  &  &  &  & \\
        MMLU  & {67.2}  & {72.5}  & {67.0}  & \underline{78.6}  & \textbf{82.9}  \\
        MMLU-Pro & {44.6}  & {49.2}  & {43.3}  & \underline{56.9}  & \textbf{71.9}  \\
        CMMLU & {64.3}  & {67.5}  & {58.8}  & \underline{75.7}  & \textbf{79.8}  \\
        CEval & {63.9}  & {66.5}  & {58.0}  & \textbf{72.7}* & \underline{72.5}  \\
        IFEval & {59.3}  & \textbf{62.5}  & \textbf{62.5}  & {61.4}  & {60.8}  \\
        \midrule
        \textit{Math} &  &  &  &  & \\
        GSM8K & {82.4}  & {89.0}  & {81.0}  & \textbf{91.3}  & \underline{91.0}  \\
        MATH  & {58.7}  & {73.5}  & {39.2}  & \underline{78.6} & \textbf{84.0}  \\
        \midrule
        \textit{Coding} &  &  &  &  & \\
        MBPP  & {70.0}  & \underline{77.8}  & {58.8}  & {72.0}  & \textbf{87.6}  \\
        HumanEval & {61.6}  & \underline{80.5}  & {55.5}  & {78.7}  & \textbf{89.0}  \\
        \midrule
        \textbf{Avg} & {61.1} & {71.0} & 58.2 & \underline{74.0} & \textbf{79.9} \\
        \bottomrule
    \end{tabular}
    }
    \vspace{2mm}    
    \label{tab:sftnew}
\end{table}

\subsection{Evaluation}

We primarily evaluate \textsc{NBDiff-7B} across three capability areas—code, math, and general knowledge—and compare its performance against several baseline models to understand relative strengths and trade-offs. We evaluate general capabilities on MMLU~\cite{mmlu}, MMLU-Pro~\cite{mmlupro}, CEVAL~\cite{eval}, CMMLU~\cite{cmmlu}, and IFEval~\cite{ifeval}; mathematical reasoning on GSM8K~\cite{gsm8k} and MATH~\cite{MATH}; and coding performance on MBPP~\cite{mbpp} and HumanEval~\cite{humaneval}.

We present the SFT (Instruct) results (\textit{i.e.} \textsc{NBDiff-7B-Instruct}) in Table~\ref{tab:sftnew}. Due to page limits, the Base version, \textsc{NBDiff-7B-Base}, is presented in Appendix~\ref{sec:otherexperiments}. \textsc{NBDiff-7B-Instruct} delivers the highest macro average (79.9) among SFT baselines, substantially outperforming SDAR-8B~\cite{sdar} and LLaDA~\cite{llada} variants~\cite{lladamoe,llada20}. On general knowledge, it sets the pace on MMLU (82.9), MMLU-Pro (71.9), and CMMLU (79.8), and ranks second on CEval (72.5) while remaining competitive on IFEval (60.8) despite ties among baselines. For math, \textsc{NBDiff-7B-Instruct} achieves good GSM8K performance (91.0) and State-of-the-Art MATH performance (84.0), indicating strong multi-step and competition-style reasoning under instruction following. In coding, it tops both MBPP (87.6) and HumanEval (89.0), narrowing and in most cases reversing the AR-favoring gap seen in some base models. Taken together, these results show that instruction tuning on a diffusion LLM not only preserves the Base model's breadth, but amplifies performance across general, math, and coding by large margins, establishing \textsc{NBDiff-7B-Instruct} as a strong, balanced SFT model in the 7B class.

\section{Conclusion}
\label{sec:conclusion}

In this work, we propose a principled adaptation framework that bridges the gap between Autoregressive (AR) and Block-Diffusion models. By reframing adaptation as a continuous interpolation--viewing AR as a Block-Diffusion model with a block size of one--we introduce the context-causal attention mechanism and an efficient parallel training recipe with auxiliary AR supervision, which maximally preserves the pre-trained knowledge of the source model. We also propose a block-size growth curriculum that smoothly transitions the model from sequential to parallel generation.

Our resulting model, \textsc{NBDiff-7B}, has achieves state-of-the-art performance among 7B-parameter diffusion models, outperforming strong baselines on math, code, and general reasoning benchmarks. These results demonstrate that expensive pre-training from scratch is not necessary to build high-quality diffusion LLMs. Instead, our method offers a compute-efficient pathway to unlock parallel generation capabilities in existing open-source AR checkpoints, potentially accelerating the deployment of faster and more flexible generative models.

\small
\bibliographystyle{plain}
\bibliography{neurips_2025}

%%%%%%%%%%%%%%%%%%%%%%%%%%%%%%%%%%%%%%%%%%%%%%%%%%%%%%%%%%%%
\section*{Contributors}

\textbf{%
  Yuchuan Tian$^{1*}$, Yuchen Liang$^{2*}$, \\Shuo Zhang$^{2}$, Yingte Shu$^{1}$, Guangwen Yang$^{2}$,  Wei He$^{1}$, Sibo Fang$^{2}$, Tianyu Guo$^{2}$, Kai Han$^{2}$, \\Chao Xu$^{1}$, Hanting Chen$^{2\dagger}$, Xinghao Chen$^{2\#}$, Yunhe Wang$^{2\#}$\\ \\}
  \small$^1$ State Key Lab of General AI, School of Intelligence Science and Technology, Peking University. \\
  \small$^2$ Huawei Technologies. \\
  \small $^*$Equal Contribution. $^\dagger$Project Lead. $^\#$Corresponding Author.

\section*{Acknowledgement}

We are very grateful to Yulong Li, Xuechun Wang, Renjie Jiang, Chen Chen and Hang Zhou for their generous help.

\newpage
\appendix
\section{Methodology Details}
\label{sec:methodologydetails}
\textbf{Attention mask for parallel training.} The naive block-diffusion recipe is data-inefficient: random cropping wastes the remaining tokens of each sequence, and only a small subset of masked tokens inside the last block contributes to the loss. Unlike AR pretraining—where every token can supervise next-token prediction—switching to next-\emph{block} prediction sharply reduces token utilization. We therefore restructure training so that all blocks provide learning signal in a single forward pass.

We seek to model all blockwise conditionals in parallel using a single transformer call. Instead of invoking the denoiser $B$ times, we concatenate a \emph{noised} view $x_t$ (partitioned into blocks) with the \emph{clean} sequence $x$:
\[
x_{\mathrm{all}} \;=\; x_t \;\oplus\; x \quad\text{(length $2L$)}.
\]
A structured attention mask $\mathbf{M}_{\mathrm{all}}\in\{0,1\}^{2L\times 2L}$ updates all token representations in one shot:
\begin{equation}
\label{eq:mask_full}
\mathbf{M}_{\mathrm{all}}
=
\begin{bmatrix}
\mathbf{M}_{\mathrm{BD}} & \mathbf{M}_{\mathrm{OBC}} \\
\mathbf{0}               & \mathbf{M}_{\mathrm{CC}}
\end{bmatrix}.
\end{equation}

Within the noised view $x_t$, attention is restricted to be block-wise (block-diagonal):
\[
[\mathbf{M}_{\mathrm{BD}}]_{ij}=
\begin{cases}
1,& \text{$i$ and $j$ are in the same block},\\
0,& \text{otherwise}.
\end{cases}
\]
From noised tokens to the clean context, we allow only \emph{earlier} clean blocks as conditioning (offset block-causal):
\[
[\mathbf{M}_{\mathrm{OBC}}]_{ij}=
\begin{cases}
1,& \text{clean position $j$ lies in a block} \\
  & \text{strictly before the block of $i$},\\
0,& \text{otherwise}.
\end{cases}
\]
Inside the clean context, we keep strict left-to-right causality (context-causal):
\[
[\mathbf{M}_{\mathrm{CC}}]_{ij}=
\begin{cases}
1,& j \le i,\\
0,& j>i.
\end{cases}
\]
The lower-left tile is zero so the clean context never reads from the noised view, matching inference-time semantics.

Let $\mathcal{B}$ index all blocks and $\mathcal{M}_t$ be the step-dependent visibility inside the noised view. Under Eq.~\eqref{eq:mask_full}, one forward pass supplies gradients for all masked tokens across all blocks:
\begin{equation}
    \label{eq:parallel_loss}
    \begin{split}
    \mathcal{L}_{\mathrm{parallel}}(\theta)
    &=
    \mathbb{E}\!\left[
    -\sum_{B\in\mathcal{B}}
    \sum_{i\in B:\,\mathcal{M}_t(i)=0} \right. \\
    &\quad \left. \log p_{\theta}\big(x_i \mid x_{t}, x;\, \mathbf{M}_{\mathrm{all}}\big)
    \right].
    \end{split}
    \end{equation}
Processing $x_t$ and $x$ jointly amortizes KV-cache construction, maximizes per-step token utilization, and empirically stabilizes training compared with randomly growing global masks. An example for $L{=}16$ and block size $b{=}4$ is illustrated in Fig.~\ref{fig:plot2}; but in reality, we use $b{=}32$.

\section{Other Experiments}
\label{sec:otherexperiments}
\paragraph{NBDiff-7B-Base.} We also measure the performance of \textsc{NBDiff-7B-Base}. The comparative results for our \textsc{NBDiff-7B-Base} against strong 7B baselines is summarized in Table~\ref{tab:base}. Apart from the introduced benchmarks, we also include BBH (BIG-Bench Hard)~\cite{bbh}, which is a curated set of particularly difficult tasks targeting abstraction, compositionality, and complex reasoning. Overall, \textsc{NBDiff-7B-Base} attains the highest macro average, surpassing Dream-v0-Base-7B and both LLaDA bases. On general knowledge, it leads on MMLU-Pro (52.7), CMMLU (76.9), CEval (75.9), and BBH (69.4), and remains competitive on MMLU (69.1, second only to Dream's 69.5). In math, \textsc{NBDiff-7B-Base} ranks first on both GSM8K (79.6) and MATH (46.0), indicating strong multi-step and competition-style reasoning. In coding, it is consistently runner-up, slightly behind Dream-v0~\cite{dream} but ahead of the LLaDA~\cite{llada} baselines. Taken together, these results show that a diffusion-style LLM can match or outperform autoregressive bases across diverse evaluations, with particularly clear gains on harder general-reasoning and Chinese benchmarks.

\begin{table}[htbp]
    \centering
    % \small
    \setlength{\tabcolsep}{4pt}
    \caption{\textbf{Comparison between \textsc{NBDiff-7B-Base} and latest base-version diffusion language models.} Our base model shows strong performance on general, math, and coding benchmarks.}
    \scalebox{0.9}{
    \begin{tabular}{lccc>{\columncolor{gray!10}}c}
    \toprule
        & \textbf{LLaDA-8B} & \textbf{LLaDA-MoE-7B} & \textbf{Dream-v0} & \textbf{Ours-7B} \\
    {Benchmark} & \textbf{Base} & \textbf{A1B-Base} & \textbf{Base-7B} & \textbf{Base} \\
    \midrule
    \textit{General} &  &  &  &  \\
    MMLU      & {65.9} & {64.6} & \textbf{{69.5}} & \underline{{70.1}} \\
    MMLU-Pro  & {41.8} & {39.2} & \underline{{48.2}} & \textbf{{59.1}} \\ 
    CMMLU     & \underline{{69.9}} & {65.7} & {60.9} & \textbf{{77.3}} \\
    CEval     & \underline{{70.5}} & {65.6} & {59.2} & \textbf{{73.0}} \\
    BBH       & {49.8} & {52.7} & \underline{{57.9}} & \textbf{{77.3}} \\
    \midrule
    \textit{Math} &  &  &  &  \\
    GSM8K     & {70.7} & {66.4} & \underline{{77.8}} & \textbf{{78.8}} \\
    MATH      & {27.3} & {36.1} & \underline{{39.6}} & \textbf{{46.0}} \\
    \midrule
    \textit{Coding} &  &  &  &  \\
    MBPP      & {38.2} & {52.4} & \textbf{{56.2}} & \underline{{55.8}} \\
    HumanEval & {33.5} & {45.7} & \textbf{{57.9}} & \underline{{50.0}} \\
    \midrule
    \textbf{Avg} & {52.0} & {54.3} & \underline{{60.1}} & \textbf{{65.3}} \\
    \bottomrule
    \end{tabular}
    }
    \vspace{2mm}    
    \label{tab:base}
\end{table}

\begin{table}[htbp]
    \centering
    \caption{\textbf{Comparing DLMs adapted from Base and SFT version of loaded weights.} Context-Causal gains a clear advantage in adaptation from AR.}
    \setlength{\tabcolsep}{2pt}
    \scalebox{0.83}{
    \begin{tabular}{lccccc}
    \toprule
    Scheme & GSM8K & MATH & HumanEval & MBPP & Avg \\
    \midrule
    \rowcolor{gray!10}\textbf{Qwen3-8B-Base} & \textbf{82.26} & \textbf{34.68} & {31.71} & \textbf{64.40} & {53.26} \\
    \textbf{Qwen3-8B}   & {81.05} & {32.24} & \textbf{42.68} & {61.60} & \textbf{54.39} \\
    
    \bottomrule
    \end{tabular}
    }
    \vspace{2mm}
    \label{tab:baseorsft}
\end{table}

\paragraph{Base vs. SFT AR weight initialization.} We attempted adapting both a pretrained Base AR checkpoint and an instruction-tuned (SFT) checkpoint into our DLM, with the comparison summarized in Table~\ref{tab:baseorsft}. Somewhat surprisingly, the Base-initialized model achieves the stronger overall balance (Though Avg 53.26 vs. 54.39 for SFT, SFT’s edge is driven largely by HumanEval’s small-set volatility; for other benchmarks, the Base model performs slightly better). We hypothesize two causes: 1. Objective alignment: the Base model is trained purely on next-token prediction, which better complements our masked-diffusion objective and auxiliary AR-loss, whereas SFT shifts the likelihood landscape toward instruction formats and response conventions; 2. Format priors: SFT injects stylistic and safety priors (headings, disclaimers, verbosity) that are beneficial for chat but act as spurious targets for diffusion.

%%%%%%%%%%%%%%%%%%%%%%%%%%%%%%%%%%%%%%%%%%%%%%%%%%%%%%%%%%%%

\end{document}